\documentclass{article}

\usepackage{PRIMEarxiv}

\usepackage[utf8]{inputenc} 
\usepackage[T1]{fontenc}    
\usepackage{hyperref}       
\usepackage{url}            
\usepackage{booktabs}       
\usepackage{amsfonts}       
\usepackage{nicefrac}       
\usepackage{microtype}      
\usepackage{lipsum}
\usepackage{fancyhdr}       
\usepackage{graphicx}       
\graphicspath{{images/}}     

\usepackage[round,sort]{natbib}
 
\usepackage[LGR,T1]{fontenc}
\usepackage[utf8]{inputenc}   

\newcommand{\textgreek}[1]{\begingroup\fontencoding{LGR}\selectfont#1\endgroup}

\title{The use of the word \textgreek{γυναικοκτονία} in Greek-speaking Twitter

}

\author{
  Aglaia Aggistrioti, Efstathia Bambili, Nikoleta Gkatzoli, Athina Kontostavlaki, Ioanna Tsounidi, Konstantinos Perifanos \\
  National and Kapodistrian University  \\
  Athens, Greece\\
  \texttt{kperifanos@phil.uoa.gr}\\
   \And
}

\begin{document}
\maketitle

\begin{abstract}
Between 2019 and 2022, Greek media attention has been attracted by a rather unusually high number of femicide cases which have been trending for several weeks up to months in the public debate and one of the contributing factors is the feedback loop between traditional media and social media. In this paper we are investigating the use of the term \textgreek{"γυναικοκτονία"} (femicide) in Greek speaking twitter. More specifically, we approach the problem from a stance detection perspective, aiming to automatically identify user position with regards to the feministic semantics of the word. We also discuss findings from an identity analysis perspective and intercorrelations with hate speech that have been identified in the collected corpus of tweets. 
\end{abstract}


\section{Introduction}

The use of the word \textgreek{"γυναικοκτονία"} (femicide) in Greek traditional and social media has attracted attention the last few years\footnote{https://femicide.gr/}\footnote{Diotima - Centre for gender rights  \& equality https://diotima.org.gr/en/}. In this paper we investigate the use of the word \textgreek{"γυναικοκτονία"} in Greek speaking twitter, examining the frequency and the context of the word appearances in a period of more than 13 years. We frame the problem from a stance detection perspective where the goal is identify speaker’s stance  and opinion expression toward a given proposition \citep{kuccuk2020stance}.

In this context, the proposition is the use of the word  \textgreek{"γυναικοκτονία"}   to describe hate crimes where the victims were women and the stance is either “positive” meaning that the users are adopting the use of the word or “negative” where users reject the use of the word. Note that the stance detection approach in this case is examining the use of the term and not potentially extreme victim blaming views with respect to the actual event.

\section{Related work}

The term \textit{femicide} is defined by the  World Health Organisation (WHO) as “violence against women comprises a wide range of acts – from verbal harassment and other forms of emotional abuse, to daily physical or sexual abuse. At the far end of the spectrum is femicide: the murder of a woman”. 

Undoubtedly, femicide is a global phenomenon. It is interesting to note that the related research tends to be followed and labeled by location. A detailed study of the social and legal perspectives of femicide in Europe are discussed in \citep{weil2018femicide}. The study word usage in social media in Kenya and South Africa is discussed in \citep{okech2021feminist}.  Vera \citep{veratext} follow a text mining approach focused in Colombia and Mexico.  A legal study and statistics of femicide cases in Milan, Italy is presented by \citep{biehler2022twenty}.

Wei \citep{weil2020two} discusses the increase of the femicide and the increase of domestic violence that has been recorded during the lockdowns in the pandemic period. Apospori \citep{alphapiosigmapiorhoeta2022acutevarepsilonmuvarphiupsilonlambdavarepsilonvarsigma} discusses gender representations of the word \textgreek{“γυναικοκτονία”} in Greek speaking social media.

\section{Data and methodology}
We have collected more than 55 thousand tweets including the word \textgreek{“γυναικοκτονία”} over a period of 13 years (2009-May 2022). 
We proceed with our analysis in two levels, the quantitative level of word analysis where we examine frequencies of words and collocations per label and the qualitative level where we emphasise on the hate speech aspect and the attribution of identities of the users producing the tweets. For the quantitative analysis we used the Antconc software with further analysis using SPSS and Microsoft Excel. 

\begin{figure}
  \caption{Frequencies}  
  \centering
  \label{fig:freq}
 \includegraphics[scale=0.75]{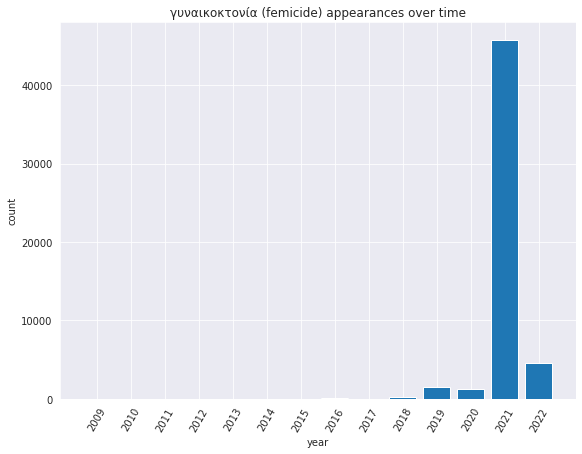}
\end{figure}

The frequency of the word per year is depicted in figure \ref{fig:freq}. We observe a rather extreme change in the use of the word in the years after 2017. It is important though to observe that the real onset seems to be the year 2016 with 57 appearances, compared to 0-9 instances until 2015.

\section{Results}
Our analysis shows that approximately 85\% of the tweets are in the positive stance.  The magnitude of the positive stance can be partially attributed to the adoption and the use of the word from mainstream media, which is also reflected in our dataset as tweets from news/media accounts that contain and adopt the term. 

Another factor that contributes to the magnitude is the dynamics of the feminist movement in social media especially during and after the lockdowns in the pandemic period. 

As for the negative use of the term, it seems that is accompanied by words such as \textgreek{«ανθρωποκτονία»} (homicide), \textgreek{«δολοφονία»} (murderer) and \textgreek{«ανδροκτονία»} (andromicide) which implies that many users equate the term femicide with homicide and murder, a behaviour that is not observed in the data where the term is used in a positive labelling. 
 
Finally, we train a machine learning classifier to essentially perform stance detection, eg identify the positive/negative aspect of the use of the word \textgreek{“γυναικοκτονία”} in a single tweet.

\begin{table}[!ht]
    \centering
   
    \caption{Percentage of hate words in positively tagged tweets}
     \label{table_1}

\begin{tabular}{lcllcl}
\textbf{Normalised frequencies} &  &  & \textbf{Normalised frequencies} &  & \tabularnewline
\cline{1-5} 
\textbf{Type of word} & \textbf{Hate speech reference} &  & \textbf{Type of word} & \textbf{Hate speech reference} & \tabularnewline
 & \textbf{Nationality} &  &  & \textbf{Ethnicity} & \tabularnewline
\cline{1-5} 
egyptian &  & 0,02 & gypsy/roma &  & 0,11\tabularnewline
albanian &  & 0,02 & taliban/tzihad/mutzahedin &  & 0,02\tabularnewline
bulgarian &  & 0,00 & eastern &  & 0,01\tabularnewline
british &  & 0,00 & Total &  & 0,13\tabularnewline
georgian &  & 0,00 &  & \textbf{Other} & \tabularnewline
\cline{4-6} 
pakistani &  & 0,01 & foreigner &  & 0,02\tabularnewline
romanian &  & 0,01 & illegal alien &  & 0,01\tabularnewline
turk &  & 0,02 & immigrant &  & 0,00\tabularnewline
\textbf{Total} &  & \textbf{0,09} & minority &  & 0,00\tabularnewline
 & \textbf{Religion} &  & refugee &  & 0,00\tabularnewline
\cline{1-2} 
islam &  & 0,02 & \textbf{Total} &  & \textbf{0,04}\tabularnewline
muslim &  & 0,08 &  &  & \tabularnewline
othoman &  & 0,00 &  &  & \tabularnewline
\textbf{Total} &  & \textbf{0,11} &  &  & \tabularnewline
\end{tabular}

\end{table}

\subsection{Instances of Hate Speech}

An interesting finding in the corpus was the presence of hate speech instances in both stances, negative and positive. Hate speech can be defined as “any communication that disparages a person or a group on the basis of some characteristic such as race, color, ethnicity, gender, sexual orientation, nationality, religion, or other characteristic.” \citep{nockleby2000hate}

The use of hate speech in social media platforms is becoming more and more frequent and intense. By exploiting the anonymity offered by social media, users insult, threaten individuals or groups of people based on some of their specific characteristics and  spreading racist and sexist stereotypes \citep{i2019multilingual,jaki2019right,halevy2022preserving}.

Not surprisingly perhaps, tweets classified to the negative stance contained a large number of hate speech (14.06\%), target mostly against Roma and Muslims \citep{mcgarry2017romaphobia, perifanos2021multimodal} as opposed to the percentage of positive stance (5.63\%), a difference that is statistically significant (x2(1)=60.18, p=0.00).  Acceptance and use of this term is expected not to coincide with the emergence of hate speech, so it is of particular interest that hate speech also appears in the positively marked tweets.

Generic references were the majority of tweets containing aggressive speech and in particular hate speech concerning ethnicity, nationality and religion. In table \ref{table_1} we present the percentage of hate-word usage after normalisation of our data.

\subsection{The role of Identity}

The emergence of hate speech can be explained by the attribution of social identities and the view of users for more favorable treatment of perpetrators by supporters of certain political ideologies. The perspective on identity assumed here is not a static and a priori given notion, but is thought of as an ongoing, never-ending process even at the individual level \citep{goffman1970strategic, benwell2006discourse}. 

Also, identity is an intersubjective phenomenon, which emphasises that identification is inherently relational, not a property of isolated individuals \citep{bucholtz2004theorizing}. In this survey, twitter users are both agents and patients of identity construction and with the use of language they manage to express and symbolise different social identities. Therefore, it is revealed that language reflects the political differentiation of the two groups, which is further triggered by the acceptance or not of the term \textgreek{γυναικοκτονία}.

From the aspect of identity, those who accept the term \textgreek{"γυναικοκτονία"} (femicide) are characterised by the opposite stance with negative characterisations related to their political beliefs (\textgreek{αριστεροί}, \textgreek{συριζαίοι} =radicals or left-wing political believers, etc.), and the broader meaning of feminism, which triggers the creation of new words with negative meaning, in order to devalue and ridicule those who belong to this group (\textgreek{φεμιναζίδια}, \textgreek{συριζοφεμινίστριες}, \textgreek{φεμιχαζές}= feminazis, femi-twats). In addition, they choose to use vocabulary that is consistent with the political beliefs of the out-group members (eg \textgreek{σύντροφοι} = comrades), but with the prospect of insulting and ridiculing the ideology of the in-group \citep{seals2018positive}.

Using these derogatory descriptions, based on their own value system, users attribute negative identities to those who accept the term feminicide and at the same time differentiate themselves. In many cases, the descriptions given to the users who are put against the term are followed by adjectives carrying negative meaning correspondingly, based on their perspective \textgreek{φαλλοκρατικός}, \textgreek{καθυστερημένος}, \textgreek{σκληρός} etc = e.g. macho, retarded, cruel, etc.). The latter emphasize - to a greater extent - the supporters’ distancing, dissatisfaction and fury, resulting in building negative identities for the respective out-group. The attitude that Twitter users adopt and the position they take influence the construction of a negative or positive identity, depending on whether they express different or common political and social beliefs respectively.

\subsection{NLP and Stance Detection}
 We provide baseline NLP models trained on the labeled dataset for stance detection and we discuss zero-shot transformer-based stance detection directions. Another focus of this work is to examine in detail the evolution of the use of the term as well as the sentiment of the use of the wor \textgreek{“γυναικοκτονία”}  and more broadly, its adoption in the language of social media, especially Twitter which is based on the free expression of each user's opinion on any topic.  

We are fine-tuning the Greek version of BERT2 and we achieve a detection performance of 0.75 f1-score.

\section{Discussion and Conclusions}
The word \textgreek{“γυναικοκτονία”} has been widely and positively adopted in Greek speaking twitter users and traditional media and news accounts. Interestingly, we identify several cases of hate speech in the corpus both from positive and negative stances. 
The vast majority of hate speech cases in this particular corpus are characterised by Romaphobia  and Islamophobia. The treatment of Roma as inferior is also evident from the high frequency of the word "gypsy" (0.11\%). 
This is reinforced by identity constructed which is also politically and socially coloured, reflected by the use of language, and more often includes the following characteristics: left or right-wing political believers, religious people or even fascists.


\bibliographystyle{unsrtnat}

\bibliography{references}

\end{document}